\documentclass[twocolumn,10pt]{article}

\usepackage[a4paper,margin=1in]{geometry}

\RequirePackage{biblatex}
\usepackage{authblk}
\usepackage{tikzducks}
\usepackage{comment}
\usepackage{subcaption}
\usepackage{bm}
\usepackage{graphicx}
\usepackage{multirow}
\usepackage{amsmath,amsfonts}
\usepackage{amsthm}
\usepackage{mathrsfs}
\usepackage[title]{appendix}
\usepackage{xcolor}
\usepackage{textcomp}
\usepackage{manyfoot}
\usepackage{booktabs}
\usepackage{listings}
\usepackage{algorithm}
\usepackage{algpseudocode}
\usepackage{soul}
\usepackage{longtable}
\usepackage{tabularx}
\usepackage{pifont,xspace}
\usepackage[table]{xcolor}

\definecolor{lightGreen}{rgb}{0.30,0.56,0}

\definecolor{eladiocolor}{HTML}{20C9BB}

\definecolor{davidecolor}{HTML}{EF330B}

\definecolor{gianlucacolor}{HTML}{093FB4}

\definecolor{evangeloscolor}{HTML}{08540C}

\newcommand{\avgRealTraffic}{y}
\newcommand{\simulator}{\hat{y}}

\newcommand{\todayDate}{June 6, 2026}

\newcommand{\realDataset}{$D_r$\xspace}
\newcommand{\augmDataset}{$D_a$\xspace}

\newcommand{\edgeSet}{\mathcal{E}}
\newcommand{\edgesWSensors}{\mathcal{E}_{s}}
\newcommand{\edge}{e}
\newcommand{\edgePrime}{e'}
\newcommand{\mappingRealVirtual}{\mathcal{Z}}
\newcommand{\score}{\mathcal{S}}
\newcommand{\candidates}{\mathcal{C}}
\newcommand{\turncount}{T}
\newcommand{\edgemetric}{M}
\newcommand{\hmin}{h_{\min}}
\newcommand{\hmax}{h_{\max}}
\newcommand{\nintervals}{K}
\newcommand{\alphaReg}{\alpha}
\newcommand{\epsExplore}{\varepsilon}
\newcommand{\cmax}{c_{\max}}
\newcommand{\deltamax}{\Delta_{\max}}

\RequirePackage{academicons}
\definecolor{orcidlogocol}{HTML}{A6CE39}
\newcommand{\myorcid}[1]{%
    \href{https://orcid.org/#1}{\textcolor{orcidlogocol}{\aiOrcid}} %
  }

\newcommand\copyrighttext{%
  \footnotesize {\centering\textbf{\color{red}(*) Preprint version. This manuscript is currently under review for possible publication in Transportation Science journal.
}}}
\newcommand\copyrightnotice{%
\begin{tikzpicture}[remember picture,overlay]
\node[anchor=south,yshift=10pt] at (current page.south) {\fbox{\parbox{\dimexpr\textwidth-\fboxsep-\fboxrule\relax}{\copyrighttext}}};
\end{tikzpicture}%
}

\title{Simulation-Driven Vehicular Traffic Data Augmentation: Extending Sensor Coverage Through Virtual Sensing}

\author[1,2]{Davide Andrea Guastella}
\author[2,3]{Eladio Montero Porras}
\author[4]{Evangelos Pournaras}
\author[2,3]{Gianluca Bontempi}

\affil[1]{Aix-Marseille University, CNRS, LIS, Marseille, France}
\affil[2]{Machine Learning Group, Université Libre de Bruxelles, Brussels, Belgium}
\affil[3]{WEL Research Institute, Wavre, Belgium}
\affil[4]{School of Computer Science, University of Leeds, UK}

\affil[ ]{\texttt{davide.guastella@lis-lab.fr}}
\affil[ ]{\texttt{eladio.montero.porras@ulb.be}}
\affil[ ]{\texttt{e.pournaras@leeds.ac.uk}}
\affil[ ]{\texttt{gianluca.bontempi@ulb.be}}

\date{}
\begin{document}
\maketitle
\copyrightnotice

\begin{abstract}Urban traffic management relies on sensor networks whose spatial coverage is limited by deployment costs and privacy regulations. Machine learning models trained on such sparse data cannot generalize to unmonitored locations and must be retrained whenever the sensor infrastructure changes. We propose a simulation-based methodology that addresses this problem by generating \emph{augmented} traffic count datasets in which each physical sensor is replaced by a \emph{virtual sensor} placed at a surrogate location in the road network. Virtual sensors are selected by a graph-search heuristic that jointly maximises vehicle-flow continuity and traffic-metric similarity between the original and surrogate locations, while enforcing a minimum spatial displacement to ensure diversity of observed traffic conditions. We validate the method on two Belgian cities: Brussels, using a calibrated model, and Namur, using synthetic models. The augmented datasets preserve the bimodal daily demand profile and the dynamics of traffic at the observed locations.
\end{abstract}

\section{Introduction}\label{sec1}
Cities worldwide are deploying data-driven systems for traffic signal control, congestion prediction, and route recommendation~\cite{mumuni_data_2022,chen_comprehensive_2024}. The performance of these systems depends on the availability of large, spatially diverse traffic count datasets. In practice, however, urban sensor networks cover only a fraction of the road network. Despite the growing adoption of traffic detectors in urban environments, their coverage is quite sparse because of the high installation and maintenance costs, particularly at suburban intersections or segments~\cite{Xing01092024}.

Consider a network monitored by sensors placed on a subset of arterial roads. The sensors sparsity creates two compounding problems. First, a predictive model trained on data from a fixed set of sensors learns patterns specific to those locations. Second, when the sensor infrastructure changes, because a detector is relocated, a new road is built, or a privacy policy mandates the removal of surveillance cameras, the collected dataset becomes partially obsolete, and the cost of recollecting representative training data can be prohibitive~\cite{10360829}. 

This paper presents a simulation-based framework that generates augmented traffic datasets from calibrated or synthetic traffic models while preserving the structure of the original sensor network. It is based on a heuristic that identifies virtual sensor locations and generates realistic traffic measurements at those locations, using turn-count data and edge-level traffic statistics derived from a calibrated or synthetic simulation model. The resulting augmented dataset preserves the structural and temporal patterns of real traffic while extending coverage to road segments not directly monitored by physical sensors.

The objective of this work is not to improve a specific prediction or control task, but rather to establish a methodology for generating realistic virtual traffic observations that can serve as an intermediate asset for a broad range of applications. However, evaluating those application-specific benefits is beyond the scope of this paper.

\begin{figure}[!ht]
    \centering
    \includegraphics[width=\linewidth]{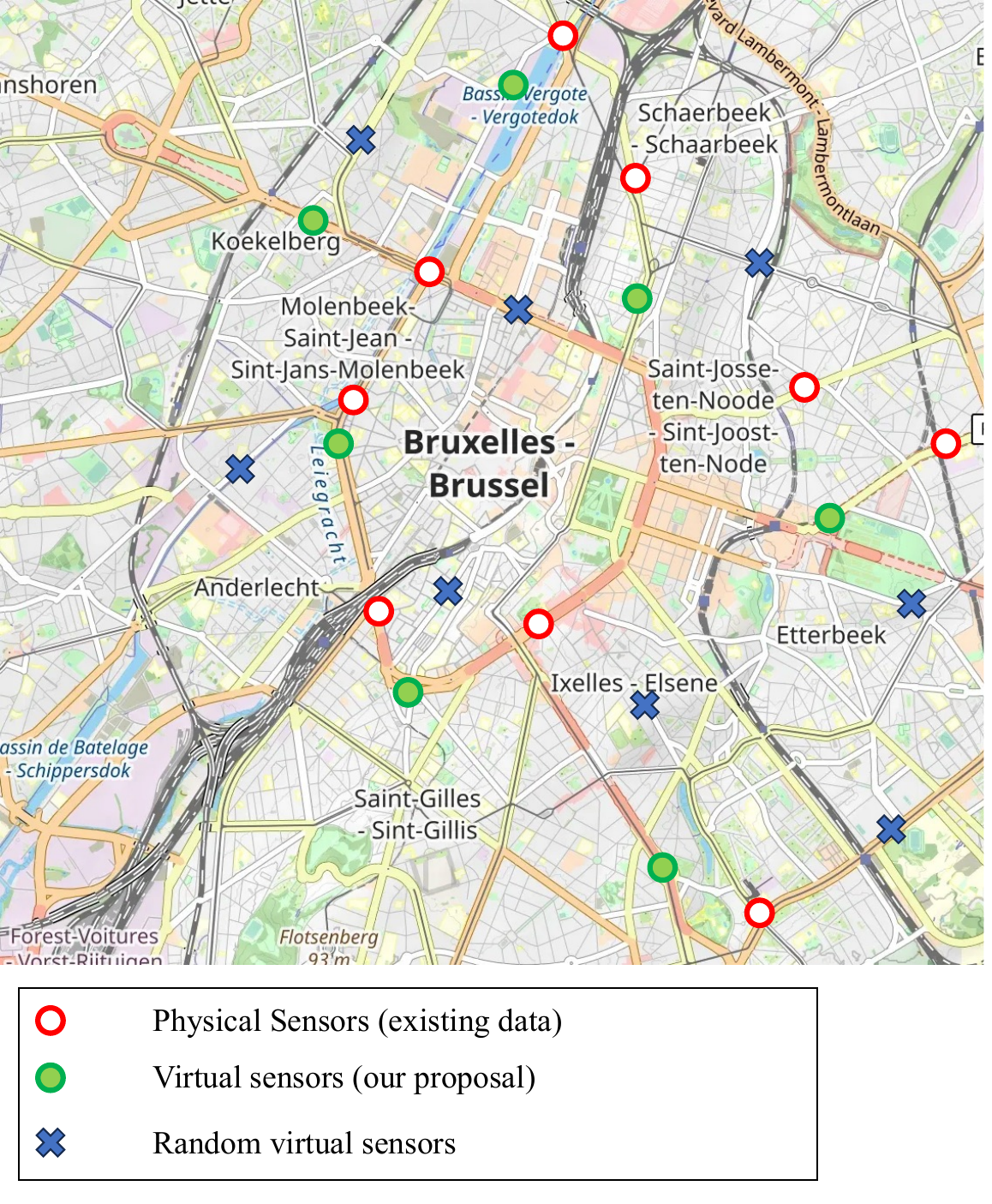}
    \caption{Example of traffic data augmentation on a part of the Brussels road network. Physical sensors (red) provide sparse coverage. Our method places virtual sensors (green) at locations with similar traffic dynamics within a limited topological distance, improving spatial coverage while preserving statistical consistency. Random placements (blue) may not capture representative traffic conditions.}
    \label{fig:case_study}
\end{figure}

To illustrate the challenges posed by sparse sensor deployments, consider a portion of the road network of Brussels, Belgium (Figure~\ref{fig:case_study}). In this area, traffic conditions are monitored by a limited number of physical sensors (red circles). These sensors are primarily installed on major arterial roads and ring roads, while many secondary and residential streets remain uninstrumented.

For many traffic modeling and management tasks, such as demand estimation, travel time prediction, or scenario analysis, data from these unmonitored locations are still required. Deploying additional sensors to fill the gaps is often infeasible due to high installation costs, limited maintenance budgets, physical constraints, or privacy concerns~\cite{ACCIAI2026132713}.

A straightforward alternative would be to place virtual sensors randomly across the network (blue crosses). However, these locations may experience traffic patterns that are very different from those of the original sensors, yielding unrealistic or uninformative data that provide limited benefit for traffic analysis tasks. Generative approaches such as GANs or LLMs learn a data distribution purely from historical observations, which is precisely the resource that is scarce or unavailable at unmonitored locations: without any prior measurements at an edge, a purely data-driven generator has nothing to condition on and cannot guarantee that its output respects the physical constraints of the road network (flow conservation, turning capacities, signal timings). 

The proposed augmentation method addresses this issue by selecting, for each existing sensor, a surrogate location within the surrounding network (within a limited number of topological hops) that exhibits similar traffic behavior in terms of volume, speed, occupancy, and travel time. Virtual sensors are then placed at these locations (green circles), ensuring a one-to-one correspondence with the original sensors. This results in an augmented dataset that preserves the statistical characteristics of the original observations while extending the spatial coverage to previously uninstrumented road segments.

The remainder of this paper is organized as follows: Section~\ref{sec:background} reviews related work on traffic data augmentation and sensor placement. Section~\ref{sec:method} presents the proposed method. Section~\ref{sec:exp} describes the experimental setup, the two case studies, the results, and the baseline comparison. Section~\ref{sec:discussion} discusses limitations and the underdetermination of the calibration problem which inherently affect the augmentation process. Finally, Section~\ref{sec:conclusion} concludes and outlines directions for future work.

\section{Background}\label{sec:background}

In many real-world settings, it is often not feasible to obtain sufficient training data to adequately train machine learning models~\cite{el_emam_practical_2020}. Data augmentation is an effective method to address this problem~\cite{mumuni_data_2022}. Generating synthetic data is a key approach to expand small or biased datasets that could contain sensitive information, particularly when data is limited~\cite{mashhadi_traffic_2025}. It has been used in fields like healthcare and finance for tasks such as diagnostic classification and fraud detection~\cite{kapp_generative_2023}. In the vehicular traffic domain, data augmentation is valuable because sensors like induction loops and cameras are often sparsely distributed and do not spatially cover the entire road network. Collecting and analyzing the stored places and times of users' presence regularly could help in inferring various important information about the users and recognizing their behaviors, social habits, customs, hobbies, workplaces, and times of travel~\cite{albouq_double_2020}. However, this has a non-negligible privacy cost for citizens.

Recent research has increasingly combined traffic sensor placement and data augmentation to address the challenges of limited network observability and sparse traffic measurements. Almutairi and Owais~\cite{almutairi_reliable_2025} propose a framework that integrates the Traffic Sensor Location Problem (TSLP), the problem of determining the optimal number and locations of sensors under budget and coverage constraints, with machine-learning-based traffic reconstruction. Their approach first identifies a minimal set of sensor locations capable of providing sufficient network coverage and then employs Stacked Sparse Auto-Encoders to infer traffic conditions on unobserved links. The resulting augmented traffic information is used to support route recommendations that minimize travel time. In the considered synthetic network, the authors report that only 21\% of road segments need to be instrumented to enable effective network-wide traffic estimation and routing.

Recent advances in the TSLP literature have increasingly focused on observability, information gain, and scalable optimization techniques. Hu and Fan~\cite{hu_sensor_2024} formulate sensor placement as a network observability problem and develop complementary algebraic and graph-theoretic approaches to maximize the observability of traffic states while minimizing the number of deployed sensors. Similarly, Yang et al.~\cite{yang_information_2024} propose an information-gradient framework that places sensors according to their expected contribution to network-wide information gain, enabling efficient deployment in large transportation systems. From an optimization perspective, Li et al.~\cite{LI202329} demonstrate that several classes of traffic sensor placement problems exhibit submodular properties, allowing near-optimal solutions to be obtained through computationally efficient greedy algorithms with theoretical performance guarantees. 

Complementing these approaches, Mashhadi et al.~\cite{mashhadi_traffic_2025} propose a methodology that leverages Variational AutoEncoders (VAE) for data augmentation in work zone traffic estimation and incorporates a custom regularized loss function to enhance model robustness. VAEs are parametric generative models that learn a low-dimensional latent space capturing essential features of the input data, which is used to generate realistic synthetic samples. The method was evaluated on a dataset of 212,000 hourly traffic volume records from Utah work zones (2016--2019). The synthetic data closely matched actual volumes, however, the nature of the input dataset could impact the performance of the data augmentation process. Notably, larger or more complex VAE architectures require more data to avoid overfitting and to learn meaningful latent space representations. In sparse sensor environments, this makes effective data augmentation more challenging.

Pham et al.~\cite{pham_finding_2025} present a data-driven method for placing traffic sensors to reconstruct traffic data. The core idea is to rank traffic links based on their importance for reconstructing traffic information from sparsely placed sensors. The method is based on two steps: (\textit{i}) Matrix Factorization with Column Pivoting, to select a subset of measurements in the original space for maximal reconstruction; (\textit{ii}) a deep reinforcement learning model, based on a Policy Gradient and a One-Dimensional Convolutional Neural Network (CNN), to learn a sensor placement policy that preserves shortest paths as the original data. A limitation of the method is its time complexity, which could affect its applicability to very large networks or when increasing the granularity of link data through more segmentations.

Existing work focuses either on reconstructing missing traffic states from sparse observations or on optimizing sensor deployment. In this work, we treat sensor observation generation as a network-level surrogate task, requiring generated data to preserve the global traffic dynamics of an existing sensor configuration rather than interpolate locally. To the best of our knowledge, this specific formulation has not been previously studied.


\section{Proposed Method}\label{sec:method}

Real traffic sensors cover only a fraction of the road network, leaving large portions of it unobserved. When a calibration method estimates origin--destination demand or route choices from sensor data, the resulting model is constrained to reproduce measurements only at instrumented locations. Sensor sparsity therefore limits both the spatial resolution and overall fidelity of calibrated models. Herein, traffic calibration is the process of adjusting the parameter of a traffic model (trajectories, number and starting time of vehicles) so that the simulator software produces output that match real-world traffic dynamics.

The proposed method addresses this limitation by exploiting traffic simulation to derive \emph{virtual} sensor readings at edges of the road network that are not covered by physical sensors. It takes as input a traffic model from which turn-count and edge-metric data are extracted. This model can be either a \emph{calibrated} model, estimated from historical sensor measurements or origin--destination matrices, or a \emph{synthetic} model, constructed from prior knowledge of the network and plausible demand patterns. Within a simulation, traffic measurements can be extracted at \emph{any} edge at no additional cost, making it possible to construct an extended dataset that mirrors the structure of a real sensor network but with broader spatial coverage. This augmented dataset is then provided as input to a calibration method. Because the augmented measurements are drawn from the same simulation that defines the reference traffic profile, the augmented model is expected to reproduce a vehicle-flow profile consistent with the one produced by the input model.

Table~\ref{tab:symbols} summarizes the symbols used throughout this section.

\begin{table}[!ht]
\small
\centering
\caption{Symbols and notation used in the proposed method.}
\label{tab:symbols}
\begin{tabularx}{\columnwidth}{|l|X|}
\hline
\textbf{Symbol} & \textbf{Description} \\ \hline
$\edgeSet$ &
  Set of all edges in the road network (with and without physical sensors).
\\ \hline
$\edgesWSensors \subseteq \edgeSet$ &
  Subset of edges equipped with physical sensors.
\\ \hline
$[\hmin,\, \hmax]$ &
  Integer interval bounding the hop distance of a virtual sensor from its paired physical sensor. Candidate locations closer than $\hmin$ hops are excluded as locally redundant; those beyond $\hmax$ hops are excluded as likely uncorrelated with the physical sensor.
\\ \hline
$d_{\mathrm{hop}}(\edge,\, \edgePrime)$ &
  Hop distance between two edges: the minimum number of directed edges traversed along any path from the downstream junction of $\edge$ to the upstream junction of $\edgePrime$. Returns $0$ when $\edge = \edgePrime$ and $+\infty$ when no directed path exists.
\\ \hline
$\turncount^{t}(\edge,\, \edgePrime)$ &
  Simulated turn-count from edge $\edge$ to adjacent downstream edge $\edgePrime$ during hourly interval $t$: the number of vehicles observed in simulation transitioning from $\edge$ to $\edgePrime$ within interval $t$.
\\ \hline
$\edgemetric^{t}(\edge)$ & Scalar traffic metric for edge $\edge$ in interval $t$ (e.g.\ average speed, travel time, or occupancy), as produced by the micro-simulation.
\\ \hline
$\nintervals$ &
  Total number of hourly intervals in the simulation period.
\\ \hline
$\mappingRealVirtual : \edgesWSensors \to \edgeSet \setminus \edgesWSensors$ &
  Injective mapping associating each physically sensored edge with a distinct virtual sensor edge. Injectivity ensures that no two physical sensors share the same virtual location, and that virtual sensors are never placed on already-sensored edges.
\\ \hline
$\candidates(t,\, \edge)$ &
  Set of candidate virtual-sensor edges at time interval $t$ for sensored edge $\edge$, produced by Algorithm~\ref{alg:get_candidates}.
\\ \hline
$\score(\edge,\, \edgePrime)$ &
  Cumulative score (accumulated over all $\nintervals$ intervals) evaluating the suitability of placing a virtual sensor on $\edgePrime$ as a surrogate for the physical sensor on $\edge$. A higher score reflects greater similarity in both turning behaviour and simulated traffic dynamics.
\\ \hline
$\alphaReg \in [0,1]$ &
  Regularization weight balancing turn-count similarity ($\alphaReg = 1$) and traffic-metric similarity ($\alphaReg = 0$) in the scoring function.
\\ \hline
$\epsExplore \in [0,1]$ &
  Exploration rate for stochastic virtual-sensor assignment: with probability $\epsExplore$ a candidate is drawn uniformly at random from the eligible pool instead of selecting the highest-scoring assignment.
\\ \hline
\end{tabularx}
\end{table}

The proposed method constructs an injective mapping $\mappingRealVirtual : \edgesWSensors \to \edgeSet \setminus \edgesWSensors$ associating each physically sensored edge in $\edgesWSensors$ with a distinct \emph{virtual sensor} placed on a distinct edge. The pairing is not arbitrary: a virtual sensor is considered a good surrogate for a physical sensor when the traffic dynamics it exhibits under simulation are similar to those observed at the sensored edge. Similarity is quantified by a regularized scoring function that jointly considers (\textit{i}) the volume of vehicles flowing between the two edges (turn-count similarity) and (\textit{ii}) the properties of the observed traffic, such as average speed or occupancy.

The method is decomposed into two procedures. The first, \textsc{GetCandidates} (Algorithm~\ref{alg:get_candidates}), is a subroutine that, given a sensored edge $\edge$ and a simulation interval $t$, returns the set of candidate edges eligible for virtual sensor placement. The second is the main TSLP heuristic (Algorithm~\ref{alg:augmentation}), which invokes \textsc{GetCandidates} for every sensored edge and every hourly interval, accumulates a pairwise score $\score(\edge, \edgePrime)$ that jointly reflects turn-count and traffic-metric similarity, and finally constructs the injective mapping $\mappingRealVirtual$ via a global greedy matching.

\subsection{Candidate identification}

For each physically sensored edge $\edge \in \edgesWSensors$ and each hourly interval $t$, \textsc{GetCandidates} identifies the edges eligible as virtual sensor locations by performing a breadth-first traversal of the directed road network graph. The search starts from $\edge$ and follows only edges along which simulated vehicles were observed, that is, edges $\edgePrime$ with $\turncount^{t}(\cdot, \edgePrime) > 0$, and collects those whose hop distance from $\edge$ falls within $[\hmin, \hmax]$. The lower bound value $\hmin$ prevents virtual sensors from being placed too close to $\edge$, where traffic conditions would be nearly identical and thus provide no additional diversity to the augmented dataset. Similarly, the upper bound $\hmax$ prevents placements so remote that the traffic dynamics at the candidate edge are likely uncorrelated with those at the physical sensor. Restricting the traversal to edges with positive turn counts has a dual purpose: it confines the search to parts of the network that actually carry simulated traffic during the interval of interest, and it implicitly encodes the direction of vehicle flow, so that virtual sensors are placed along plausible downstream routes rather than in topologically reachable but traffic-disconnected locations. It is worth noting that hop distance is a purely topological measure that does not account for geometric edge lengths or travel times; the scoring function described in the following compensates for this by penalizing candidates whose simulated traffic properties deviate from those at the physical sensor.

\begin{algorithm}[!ht]
\caption{\textsc{GetCandidates}: BFS-based candidate edge search}
\label{alg:get_candidates}
\begin{algorithmic}[1]
\Require Sensored edge $\edge$; hop-distance bounds $\hmin$, $\hmax$;
         turn-count dictionary $\turncount^{t}$
\Ensure  Candidate set $\candidates(t,\, \edge) \subset
         \edgeSet \setminus \edgesWSensors$
\State $\mathcal{Q} \gets \{(\edge,\; 0)\}$,\quad
       $\mathcal{V} \gets \{\edge\}$,\quad
       $\candidates \gets \emptyset$
       \label{line:gc_init}
\While{$\mathcal{Q} \neq \emptyset$}
    \label{line:gc_bfs}
    \State $(e_{\mathrm{cur}},\, h) \gets \mathcal{Q}.\mathrm{dequeue}()$
    \ForAll{$\edgePrime \notin \mathcal{V}$ \textbf{with}
            $\turncount^{t}(e_{\mathrm{cur}},\, \edgePrime) > 0$}
        \label{line:gc_neighbors}
        \State $h' \gets h + 1$
        \If{$\hmin \leq h' \leq \hmax$}
            \State $\candidates \gets \candidates \cup \{\edgePrime\}$
            \label{line:gc_add}
        \EndIf
        \If{$h' < \hmax$}
            \label{line:gc_enqueue}
            \State $\mathcal{V} \gets \mathcal{V} \cup \{\edgePrime\}$,\quad
                   $\mathcal{Q}.\mathrm{enqueue}(\edgePrime,\, h')$
        \EndIf
    \EndFor
\EndWhile
\State \Return $\candidates$
\label{line:gc_return}
\end{algorithmic}
\end{algorithm}

Algorithm~\ref{alg:get_candidates} proceeds as follows. The BFS frontier $\mathcal{Q}$, the visited set $\mathcal{V}$, and the candidate set $\candidates$ are initialized at line~\ref{line:gc_init}: the frontier contains only the source edge $\edge$ at hop distance $0$, and both $\mathcal{V}$ and $\candidates$ are set accordingly. The BFS loop (line~\ref{line:gc_bfs}) iterates until the frontier is exhausted. At each step, the algorithm dequeues the current edge $e_{\mathrm{cur}}$ together with its hop distance $h$ from $\edge$, then examines all unvisited downstream neighbors $\edgePrime$ with a strictly positive turn count $\turncount^{t}(e_{\mathrm{cur}}, \edgePrime) > 0$ (line~\ref{line:gc_neighbors}), ensuring that only edges along active simulated traffic flows are considered. The tentative hop distance $h' = h + 1$ is computed for each such neighbor. If $h'$ lies within the prescribed interval $[\hmin, \hmax]$, the edge $\edgePrime$ is added to the candidate set (line~\ref{line:gc_add}). If $h'$ is still strictly below $\hmax$, the neighbor is additionally marked as visited and enqueued for further exploration (line~\ref{line:gc_enqueue}); this prevents re-visiting edges and avoids exploring beyond the prescribed depth. Edges at exactly $\hmax$ hops are collected as candidates but \emph{not} enqueued, correctly terminating the search at the outer boundary. The subroutine returns the candidate set $\candidates$ at line~\ref{line:gc_return}.

\subsection{Scoring function}

Not all candidate edges are equally suitable as virtual sensor locations. A candidate $\edgePrime \in \candidates(t, \edge)$ is a good surrogate for the physical sensor at $\edge$ if the traffic it observes under simulation is similar to what the physical sensor records. We quantify this notion via a regularized scoring function that combines two complementary measures, both derived from the simulation.

The first measure is \emph{turn-count similarity}. The turn count $\turncount^{t}(\edge, \edgePrime)$ measures how many simulated vehicles transit from edge $\edge$ toward $\edgePrime$ during interval $t$. A high value indicates strong vehicle-flow continuity between the two edges: traffic observed at the physical sensor is, to a large extent, the same traffic that subsequently passes through the candidate. Conversely, a low value suggests that the route population diverges between the two locations, weakening the statistical link between their respective counts. Because $\edgePrime$ does not need to be a direct neighbor of $\edge$, the relevant turn count is traced along the shortest path identified during the BFS, accumulating flow from $\edge$ to $\edgePrime$.

The second measure is \emph{traffic-metric similarity}. Turn-count continuity alone does not capture whether the two edges share comparable traffic conditions. Two edges may lie along the same route but differ substantially in, e.g., average speed or occupancy if one is upstream of a signalized intersection and the other is on a free-flow segment. Penalizing large differences $|\edgemetric^{t}(\edge) - \edgemetric^{t}(\edgePrime)|$ steers the selection toward candidates that exhibit traffic dynamics consistent with those at the physical sensor.

Both terms, turn-count and traffic-metric similarities, are normalized by $\alpha \in [0, 1]$. 
Let $\cmax$ be the maximum turn-count value for an edge $e$, and $\deltamax$ the maximum absolute difference between traffic-metric similarity at the edges $e$ (sensored) and $e'$ (candidate edge), calculated respectively as follow:

\begin{align}
  \cmax     &= \max_{\edgePrime \in \candidates(t,\edge)}\;
               \turncount^{t}(\edge,\, \edgePrime) ,
  \label{eq:cmax}\\[4pt]
  \deltamax &= \max_{\edgePrime \in \candidates(t,\edge)}\;
               \bigl|\edgemetric^{t}(\edge) - \edgemetric^{t}(\edgePrime)\bigr| .
  \label{eq:deltamax}
\end{align}

The score for each time interval $t$ is accumulated as

\begin{equation}
\resizebox{.5\linewidth}{!}{$
  \score(\edge,\, \edgePrime)
  \;\mathrel{+}=\;
  \alphaReg
    \cdot \frac{\turncount^{t}(\edge,\, \edgePrime)}{\cmax}
  \;-\;
  (1 - \alphaReg)
    \cdot \frac{|\edgemetric^{t}(\edge) - \edgemetric^{t}(\edgePrime)|}{\deltamax}
$}
\label{eq:score}
\end{equation}

Setting $\alphaReg = 1$ selects virtual sensors based exclusively on vehicle-flow continuity; $\alphaReg = 0$ selects them based exclusively on traffic-property similarity; intermediate values balance both objectives. When $\cmax = 0$ or $\deltamax = 0$ for a given $(t, \edge)$ pair (e.g.\ because all candidates share the same metric value) the corresponding term is set to zero to avoid division by zero.

The objective of the proposed augmentation process is not to maximize diversity of observations, but to generate plausible substitutes for existing sensors. Consequently, the desired candidate location is not the most informative edge in the network, but rather the edge whose traffic dynamics remain sufficiently similar to those observed at the physical sensor while still introducing spatial displacement.

It is important to note that the proposed scoring function balances similarity and diversity. Similarity ensures that the augmented observations remain consistent with the traffic patterns of the original sensors, while diversity encourages the placement of surrogate sensors on distinct road segments, increasing the spatial coverage of the augmented dataset.

\subsection{Virtual sensor assignment}

After scoring, the algorithm constructs the injective mapping $\mappingRealVirtual$ associating each physical sensor with a unique virtual edge. We adopt a greedy matching strategy: all valid (physical sensor, candidate) pairs, where each candidate is an edge $\edge \in \candidates \setminus \edgesWSensors$, are ranked in descending order of their time-averaged score over all time intervals.

Each pair $(\edge, \edgePrime)$ is accepted if and only if neither $\edge$ nor $\edgePrime$ has already been claimed by an earlier assignment. This greedy procedure approximates maximum-weight bipartite matching and ensures that the globally highest-affinity pairings are considered regardless of spatial sensor placement. In order to support stochastic exploration, we replace with probability $\epsExplore$ the ranked list by a uniformly random permutation of all valid pairs before the same claim mechanism is applied.

Sensors for which no uncontested candidate remains receive a null virtual location and are excluded from augmentation.

\begin{algorithm}[!ht]
\small
\caption{\textsc{AugmentDataset}: proposed heuristic for data augmentation}
\label{alg:augmentation}
\begin{algorithmic}[1]
\Require Sensored edge set $\edgesWSensors$; hop bounds $\hmin$, $\hmax$;
         turn-count dictionaries $\{\turncount^{t}\}_{t=1}^{\nintervals}$;
         scalar metric $\edgemetric^{t}(\edge)$ for all $\edge$, $t$;
         regularization weight $\alphaReg$; exploration rate $\epsExplore$
\Ensure  Injective mapping
         $\mappingRealVirtual : \edgesWSensors \to \edgeSet \setminus \edgesWSensors$

\State $\score(\edge,\, \edgePrime) \gets 0 \quad \forall\, \edge \in \edgesWSensors,\;
       \edgePrime \in \edgeSet \setminus \edgesWSensors$
\State $\candidates_{\cup}(\edge) \gets \emptyset \quad \forall\, \edge \in \edgesWSensors$
       \label{line:init}

\ForAll{$t \in \{1, \ldots, \nintervals\}$}
    \label{line:outer_loop}
    \ForAll{$\edge \in \edgesWSensors$}
        \label{line:inner_loop}
        \State $\candidates(t,\, \edge) \gets
               \textsc{GetCandidates}(\edge,\, \hmin,\, \hmax,\, \turncount^{t})$
               \label{line:get_candidates}
        \If{$\candidates(t,\, \edge) \neq \emptyset$}
            \label{line:nonempty}
            \State $\cmax \gets \max_{\edgePrime \in \candidates(t,\edge)}\;
                   \turncount^{t}(\edge,\, \edgePrime)$
                   \label{line:cmax}
            \State $\deltamax \gets \max_{\edgePrime \in \candidates(t,\edge)}\;
                   |\edgemetric^{t}(\edge) - \edgemetric^{t}(\edgePrime)|$
                   \label{line:dmax}
            \ForAll{$\edgePrime \in \candidates(t,\, \edge)$}
                \label{line:score_loop}
                \State $\score(\edge,\, \edgePrime) \mathrel{+}=\;
                       \alphaReg \cdot \dfrac{\turncount^{t}(\edge,\, \edgePrime)}{\cmax}
                       - (1-\alphaReg) \cdot
                         \dfrac{|\edgemetric^{t}(\edge) - \edgemetric^{t}(\edgePrime)|}
                               {\deltamax}$
                       \label{line:score}
            \EndFor
            \State $\candidates_{\cup}(\edge) \gets
                   \candidates_{\cup}(\edge) \cup \candidates(t,\, \edge)$
                   \label{line:union}
        \EndIf
    \EndFor
\EndFor

\State \textit{pool} $\gets \bigl\{\bigl(\score(\edge,\edgePrime)/\max_{e'' \in
       \candidates_{\cup}(\edge)} \score(\edge,e''),\;
       \edge,\; \edgePrime\bigr)
       \;\big|\; \edge \in \edgesWSensors,\;
       \edgePrime \in \candidates_{\cup}(\edge) \setminus \edgesWSensors\bigr\}$
       \label{line:pool}

\If{$\textsc{rand}() < \epsExplore$}
    \label{line:explore}
    \State Shuffle \textit{pool} uniformly at random
\Else
    \State Sort \textit{pool} in descending order of score
    \label{line:sort}
\EndIf

\State $\mathrm{free}_{s} \gets \edgesWSensors$,\quad
       $\mathrm{free}_{v} \gets \edgeSet \setminus \edgesWSensors$
       \label{line:free_sets}

\ForAll{$(\cdot,\; \edge,\; \edgePrime) \in \textit{pool}$}
    \label{line:greedy_loop}
    \If{$\edge \in \mathrm{free}_{s}$
        \textbf{ and } $\edgePrime \in \mathrm{free}_{v}$}
        \State $\mappingRealVirtual(\edge) \gets \edgePrime$
               \label{line:claim}
        \State $\mathrm{free}_{s} \gets \mathrm{free}_{s} \setminus \{\edge\}$,\quad
               $\mathrm{free}_{v} \gets \mathrm{free}_{v} \setminus \{\edgePrime\}$
    \EndIf
\EndFor

\ForAll{$\edge \in \mathrm{free}_{s}$}
    \label{line:null}
    \State $\mappingRealVirtual(\edge) \gets \texttt{null}$
\EndFor

\State \Return $\mappingRealVirtual$
\label{line:return}
\end{algorithmic}
\end{algorithm}
 
Algorithm~\ref{alg:augmentation} proceeds as follows. All scores and per-edge candidate union sets are initialized to zero and empty, respectively, at line~\ref{line:init}. The outer loop (line~\ref{line:outer_loop}) iterates over all $\nintervals$ hourly intervals, and the inner loop (line~\ref{line:inner_loop}) iterates over all sensored edges, so that every $(t, \edge)$ pair is processed exactly once. At line~\ref{line:get_candidates}, \textsc{GetCandidates} is invoked to retrieve the interval-specific candidate set $\candidates(t, \edge)$. The set of reachable candidates can be empty for certain time intervals, in this case the algorithm cannot associate any candidate edge to the sensored one, and the entire block is skipped (line~\ref{line:nonempty}). This may happen, for instance, when no turn-count information is available for the candidate edges. When candidates exist, two normalization constants are computed: $\cmax$ (line~\ref{line:cmax}) is the maximum turn count observed from $\edge$ toward any candidate in the current interval, and $\deltamax$ (line~\ref{line:dmax}) is the maximum absolute difference in the scalar traffic metric between $\edge$ and any candidate. These constants scale both scoring components to the interval $[0,1]$ so that $\alphaReg$ acts as a trade-off regardless of the absolute magnitudes of the raw values. The inner loop over candidates (line~\ref{line:score_loop}) then assigns $\score(\edge, \edgePrime)$ for each candidate $\edgePrime$ by Equation~\eqref{eq:score} (line~\ref{line:score}): the turn-count term rewards flow continuity between $\edge$ and $\edgePrime$, while the metric-dissimilarity term penalizes candidates whose simulated traffic conditions deviate from those at the physical sensor. At line~\ref{line:union}, the global candidate pool for $\edge$ is extended with the candidates found in the current interval; this union is used in the assignment phase so that candidates visible in any interval remain eligible for selection.

After the scoring pass, line~\ref{line:pool} builds the flat assignment pool as the set of all (\texttt{normalized score}, \texttt{sensor edge}, \texttt{candidate edge}) triplets, restricting candidates to those not already in $\edgesWSensors$ to enforce the injective constraint. The score of each pair is normalized by the maximum score achieved across its own candidate pool $\candidates_{\cup}(\edge)$, ensuring that the greedy ranking reflects relative affinity within each sensor's reachable set rather than absolute score magnitudes that are incomparable across sensors with different pool sizes.

The pool is either shuffled uniformly at random with probability $\epsExplore$ (line~\ref{line:explore}), supporting stochastic exploration, or sorted in descending order of time-averaged score (line~\ref{line:sort}), implementing the greedy strategy. Two sets tracking unmatched sensors and unclaimed virtual edges are initialized at line~\ref{line:free_sets}. The greedy matching loop (line~\ref{line:greedy_loop}) scans the pool from highest to lowest score: a pair $(\edge, \edgePrime)$ is accepted and recorded in $\mappingRealVirtual$ (line~\ref{line:claim}) only when both $\edge$ and $\edgePrime$ are still unmatched, whereupon both are immediately removed from their respective free sets. This single-pass procedure guarantees injectivity by construction: once a virtual edge has been claimed it cannot be assigned to any subsequent sensor, and once a sensor has been matched it is skipped for all remaining pool entries. Sensors that exhaust the pool without finding an uncontested candidate are assigned a \texttt{null} virtual location at line~\ref{line:null}, and the complete mapping is returned at line~\ref{line:return}.

\subsection{Dataset augmentation}

Once the mapping $\mappingRealVirtual$ is evaluated, the augmented dataset is constructed by replacing, for each physical sensor, its real vehicle count with the simulated count observed at the paired virtual edge. Formally, for sensor $s$ installed on edge $\edge$, the augmented count at interval $t$ is

\begin{equation}
  \hat{c}^{t}(s)
  = N^{t}\!\bigl(\mappingRealVirtual(\edge)\bigr) ,
  \label{eq:augcount}
\end{equation}

\noindent where $N^{t}(e')$ denotes the number of vehicles observed at virtual edge $e'$ during time interval $t$. Note that this vehicle-count quantity is distinct from the scalar traffic metric $\edgemetric^{t}(\edge)$ used in the scoring function (Equation~\eqref{eq:score}), which may represent speed, occupancy, or travel time. If $\mappingRealVirtual(\edge)$ is undefined because no valid candidate was found, the real sensor count is carried forward unchanged, ensuring the augmented dataset never contains fewer observations than the original one. The resulting augmented dataset shares the same structure as the input dataset.

\section{Experimental Results}\label{sec:exp}

We present in this section the results of the proposed methodology for augmenting traffic count data using the cities of Brussels and Namur (Belgium) as a use case. First, we present the simulation tool, the evaluation metrics, and the mobility scenarios. We then describe the results obtained by the proposed method. We focus on the comparison of traffic counts between the ground-truth and the augmented dataset, the spatiotemporal error structure, and the preservation of network-wide traffic dynamics.

\subsection{Simulation Tool}

We use the open-source traffic simulator SUMO~\cite{lopez_microscopic_2018}. SUMO can be used for microscopic simulation, where each vehicle and its dynamics are modeled individually, and mesoscopic simulation, where the movements of vehicles are modeled with queues and the traffic at intersections is modeled using a coarse model. We configure SUMO with a simplified microscopic model. This is done by partially simulating the behavior of vehicles in the intersections: vehicles are still subject to right-of-way rules (waiting at traffic lights and minor roads), but they will appear instantly on the other side of the intersection after passing the stop line. The vehicles cannot block the intersection, wait within the intersection for left turns, nor collide at the intersection\footnote{\url{https://sumo.dlr.de/docs/Simulation/Intersections.html\#internal_links}. Last visited: \todayDate}. 

\subsection{Evaluation Metrics}

To assess the accuracy of the augmented dataset, we evaluate the following
metrics to compare the traffic counts in \augmDataset to the ground-truth
\realDataset:

\begin{itemize}

\item \textbf{Mean Absolute Error (MAE)}: the average absolute difference
between augmented and real vehicle counts per sensor--hour pair:
\begin{equation}
  \textnormal{MAE}(\hat{y},y) = \frac{1}{n} \sum_{i=1}^{n} |y_i - \hat{y}_i|
\end{equation}

\noindent where $n$ is the number of sensor--hour observations, and $\simulator$ and
$\avgRealTraffic$ are the augmented and real traffic counts respectively.

\item \textbf{Root Mean Square Error (RMSE)}: the square root of the mean
squared difference between augmented and real counts, penalizing large
deviations more heavily than MAE:
\begin{equation}
  \textnormal{RMSE}(\hat{y},y) =
    \sqrt{\frac{1}{n}\sum_{i=1}^{n}(y_i - \hat{y}_i)^2}
\end{equation}

\noindent where $n$ is the number of sensor--hour observations, and $\simulator$ and
$\avgRealTraffic$ are the augmented and real traffic counts respectively.

\end{itemize}

MAE and RMSE capture complementary aspects of augmentation quality. MAE measures the average absolute deviation per sensor--hour pair and is robust to occasional large mismatches, whereas RMSE penalizes large individual errors disproportionately by squaring residuals before averaging. Low MAE and high RMSE suggest that the augmented data is accurate on average but occasionally produces large outliers, likely from a small number of mismatched virtual sensors.

\subsection{Mobility Scenarios}

\begin{figure*}[!ht]
    \centering
    \begin{subfigure}[b]{.49\textwidth}
        \centering
        \includegraphics[width=\textwidth]{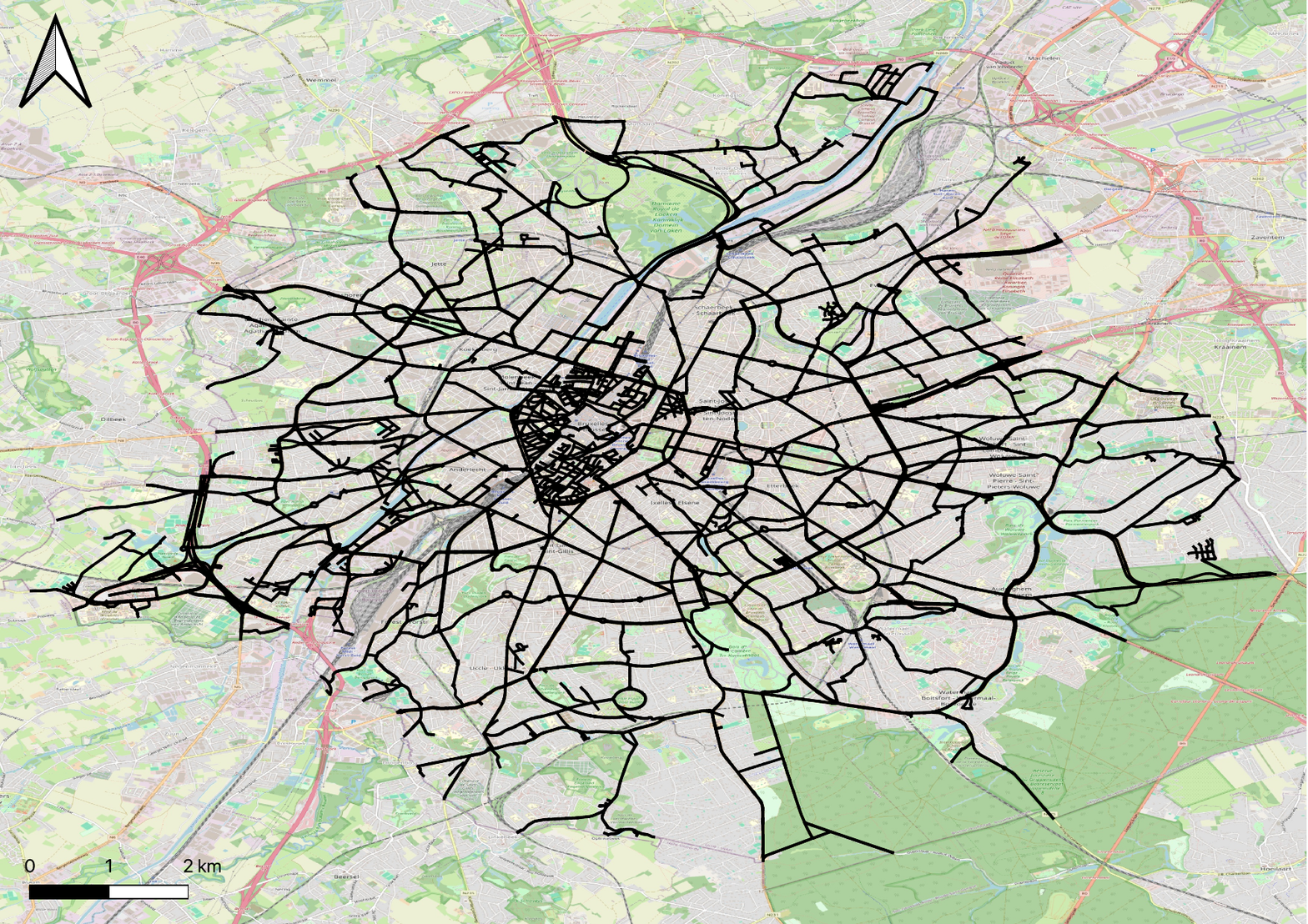}
        \caption{Brussels}
        \label{fig:bxl_net}
    \end{subfigure}
    \hfill
    \begin{subfigure}[b]{.49\textwidth}
        \centering
        \includegraphics[width=\textwidth]{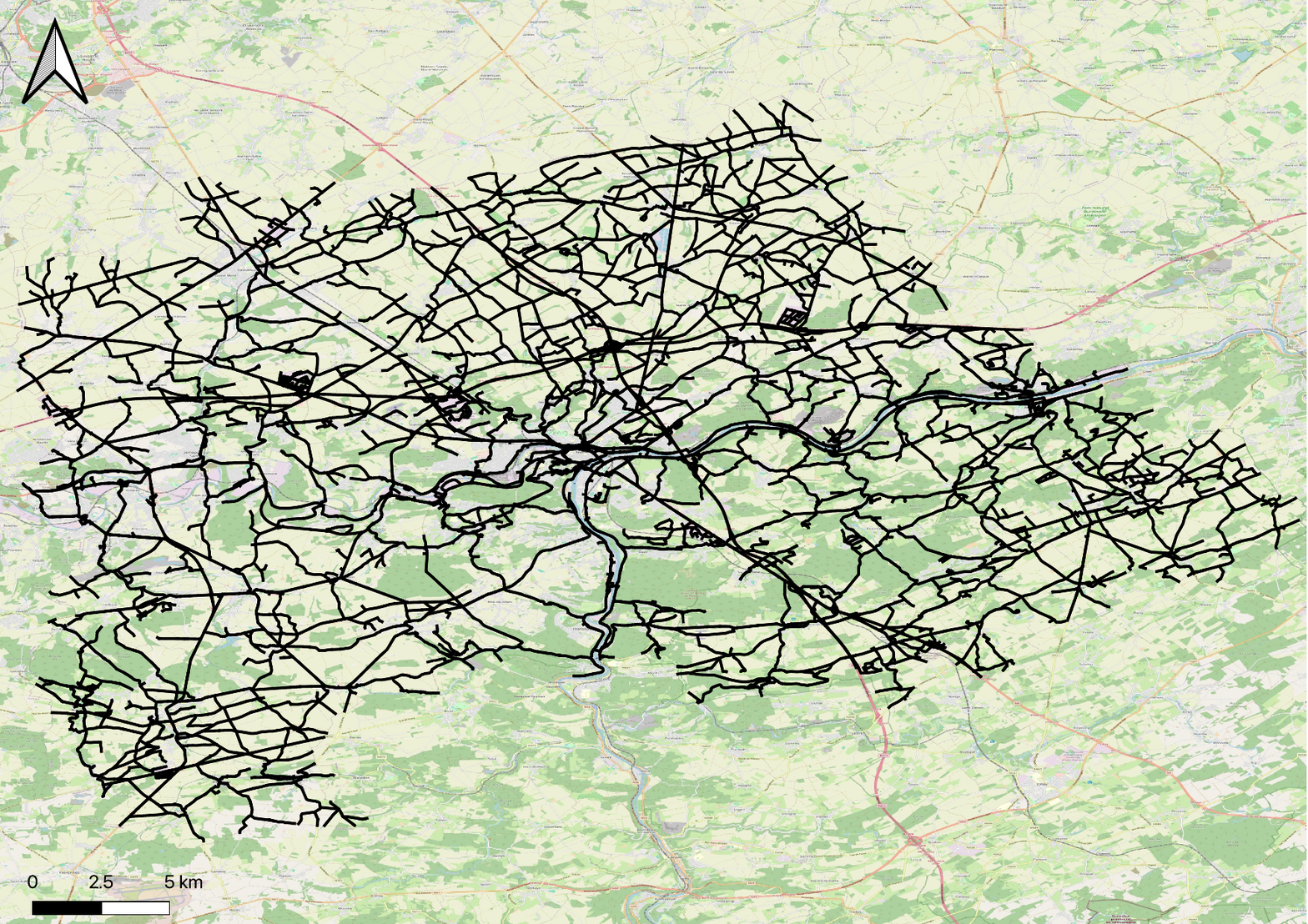}
        \caption{Namur}
        \label{fig:namur_net}
    \end{subfigure}
    \caption{Road networks of the two case-study Belgian cities. Only edges accessible to private vehicles are shown.}
    \label{fig:road_nets}
\end{figure*}

We evaluate the proposed method on the Belgian cities of Brussels and Namur. For both cities, we gather the necessary information concerning the topology of the cities from OpenStreetMap (OSM), and convert the OSM file to a format compatible with the simulation tool and filter the output road network to include only the vehicular road network. Figure~\ref{fig:road_nets} shows the road networks for the considered cities.

\subsubsection{Calibrated Traffic Models}\label{sec:result_calibmodel}

Traffic calibration from count data involves adjusting a traffic model to minimize the discrepancy between observed and simulated vehicle counts. This process ensures the model accurately reflects real-world conditions by comparing vehicle counts from physical sensors or manual observations with simulated outputs~\cite{liu_calibration_2025}.

The way the calibration method operates to estimate traffic flow has a measurable effect on the accuracy of the resulting augmented traffic model: a realistic input model yields augmented data that accurately reflects the spatial distribution of traffic demand. This highlights the importance of input model quality as a prerequisite for accurate data augmentation. In this work, we employ two calibration techniques to generate input models used to augment the traffic datasets. The first is a calibration technique established in our previous research~\cite{guastella_calibration_2025}. The calibrator takes as input a set of traffic counts and corresponding sensor locations. It begins by creating a set of region-level routes (a region can be related to a neighborhood, or administrative boundaries), then it creates link-level paths passing through the regions and through a set of sensors. The method is iterated until the simulation of the estimated model produces traffic counts that are close to the real ones. The output is a set of vehicles, each defined by a starting time and an ordered sequence of edges representing its route. The second method employs a hierarchical optimization framework that reconstructs time-varying traffic patterns from sparse data by balancing flows at a regional level before refining them into granular, edge-level trajectories~\cite{simfreecalib}. We refer to these two calibration strategies as \emph{simfree} and \emph{simbased} respectively throughout the experiments.

We use data collected from real sensor devices in Brussels (Belgium). The data has been provided by Brussels Mobility\footnote{The public administration of the Brussels-Capital Region responsible for managing and developing infrastructure, services, and strategies related to all forms of mobility within the region}. The data is collected over 24-hour periods for one day (April 3, 2024), with traffic data collected at hourly intervals. The dataset includes vehicle counts collected from 369 sensors.

\subsubsection{Synthetic Traffic Model}

Due to the unavailability of empirical traffic data, we developed synthetic models for the city of Namur (Belgium). We implemented a stochastic traffic generation procedure that distributes a fixed volume of vehicles over a 24-hour interval to replicate typical urban mobility patterns. The process begins by constructing regional-level paths across the network~\cite{guastella_calibration_2025}. Vehicle departures are then allocated based on an hourly probability vector that defines the daily demand profile. This profile incorporates a bimodal distribution with two distinct peaks, typically occurring during morning and evening rush hours (herein we assume at 8AM and 5PM), to ensure the synthetic data reflects realistic mobility patterns.

We used different volumes of vehicles, and for each we executed 30 independent runs of the scenario generation procedure. This repetition ensures that the reported results are not sensitive to a particular random realization of the synthetic demand. Each resulting traffic model is then evaluated independently against the proposed augmentation method.

\subsection{Experimental Results (Brussels Scenario)}

The experimental methodology is as follows. Starting from either a calibrated model (Brussels) or a synthetic model (Namur), we run the SUMO simulator to extract vehicle counts at edges where no physical sensor is deployed, producing an augmented dataset \augmDataset. We then compare \augmDataset against the ground-truth sensor records \realDataset using the metrics defined in Section~\ref{sec:exp}. Three hyperparameters are varied systematically: the regularization weight $\alphaReg \in \{0.0, 0.2, 0.4, 0.6, 0.8, 1.0\}$, the simulation metric used in the scoring function (occupancy, speed, travel time), and the hop-distance interval $[\hmin, \hmax]$. We evaluate three hop-range configurations:

\begin{itemize}
  \item \textbf{[1,\,5]}: restricts virtual sensors to local neighborhoods, maximizing flow continuity between physical and virtual locations.
  \item \textbf{[1,\,10]}: provides a wider candidate pool, maximizing the probability that every sensor receives a non-null assignment while still allowing the scoring function to select the best candidate.
  \item \textbf{[3,\,10]}: excludes immediate neighbors, useful for isolating the contribution of near candidates (hops 1--2) when compared to [1,\,10].
\end{itemize}

\paragraph{Hop-range sensitivity.}

Table~\ref{tab:rmse_hop_ranges} reports the RMSE and its standard deviation, averaged over all $(\alphaReg, \text{metric})$ combinations, for the three hop ranges and both calibration strategies in the Brussels scenario. The table also reports the average hop count at which virtual sensors are placed.

\begin{table}[!ht]
\centering
\caption{Augmentation error by hop range, averaged over all $(\alphaReg,\,\text{metric})$ combinations. Lower RMSE and standard deviation indicate better virtual sensors placement.}
\label{tab:rmse_hop_ranges}
\setlength{\tabcolsep}{2pt}
\renewcommand{\arraystretch}{1.15}
\small
\begin{tabular}{l r r r r r r}
\toprule
Hop Range
  & \multicolumn{3}{c}{\textbf{Simfree}}
  & \multicolumn{3}{c}{\textbf{Simbased}} \\
\cmidrule(lr){2-4}\cmidrule(lr){5-7}
  & Avg \#Hops & RMSE & STD
  & Avg \#Hops & RMSE & STD \\
\midrule
\textbf{[1,5]}  & 2.72 &  70.34 & 12.42 & 2.81 &  96.81 & 18.50 \\
\textbf{[1,10]} & 5.22 &  96.87 & 20.74 & 5.15 & 131.19 & 11.03 \\
\textbf{[3,10]} & 6.37 & 101.46 & 19.27 & 6.30 & 138.83 & 12.70 \\
\bottomrule
\end{tabular}
\end{table}

Across all configurations, the simfree strategy consistently achieves lower RMSE than simbased, and both strategies reach their minimum error under the hop range $[1,\,5]$. Widening the window to $[1,\,10]$ or enforcing a minimum gap with $[3,\,10]$ increases RMSE monotonically for both strategies, and for simfree also inflates the standard deviation, indicating that larger windows introduce candidates that are less correlated with the physical sensors they replace. This trend is reflected in the average hop count: even when the window allows up to ten hops, the scoring function selects virtual sensors within three to six hops on average, revealing that the highest-scoring candidates are consistently those closest to the physical sensor. The benefit of a wider pool is therefore marginal: the scoring function would select the same near candidates regardless, while a wider window only adds low-quality options that occasionally win the greedy matching when the best candidates are already claimed. For all subsequent analyses, we adopt the hop range $[1,\,5]$ as the recommended configuration.

\paragraph{Sensitivity to $\alphaReg$ and simulation metric.}

Table~\ref{tab:augmentation_results} reports MAE, RMSE, and the average hop count for the hop range $[1,\,5]$, grouped by $\alphaReg$ and simulation metric for both calibration strategies.

\begin{table}[!ht]
\centering
\caption{Augmentation error grouped by $\alphaReg$ and simulation metric for hop range $[1,\,5]$. \textbf{Bold} rows indicate the best configuration per strategy.}
\label{tab:augmentation_results}
\setlength{\tabcolsep}{3pt}
\renewcommand{\arraystretch}{1.1}
\small
\resizebox{\columnwidth}{!}{%
\begin{tabular}{llrrr rrrr}
\toprule
$\alphaReg$ & Metric
  & \multicolumn{3}{c}{\textbf{Simfree}}
  & \multicolumn{3}{c}{\textbf{Simbased}} \\
\cmidrule(lr){3-5}\cmidrule(lr){6-8}
& & MAE & RMSE & Avg \#Hops & MAE & RMSE & Avg \#Hops \\
\midrule
0.0 & occupancy  & 22.7 &  59.4 & 3.00 & 43.4 & 118.5 & 3.06 \\
    & speed      & 22.7 &  59.4 & 3.00 & 43.4 & 118.5 & 3.06 \\
    & traveltime & 22.5 &  55.0 & 3.23 & 41.5 & 117.2 & 3.11 \\
\textbf{0.2} & \textbf{occupancy}
              & \textbf{14.6} & \textbf{50.1} & \textbf{1.86}
              & \textbf{14.5} & \textbf{58.2} & \textbf{1.92} \\
    & speed      & 17.4 &  53.5 & 2.24 & 20.7 &  71.0 & 2.26 \\
    & traveltime & 27.3 &  77.3 & 3.07 & 32.8 & 103.1 & 3.15 \\
0.4 & occupancy  & 17.9 &  58.6 & 2.04 & 44.5 & 124.5 & 3.25 \\
    & speed      & 20.9 &  70.7 & 2.28 & 20.0 &  73.2 & 2.20 \\
    & traveltime & 24.7 &  70.0 & 3.05 & 32.8 & 102.3 & 3.11 \\
0.6 & occupancy  & 26.5 &  78.7 & 2.30 & 23.2 &  86.8 & 2.22 \\
    & speed      & 26.9 &  82.0 & 2.51 & 25.5 &  86.2 & 2.37 \\
    & traveltime & 28.0 &  81.6 & 3.13 & 29.6 &  94.0 & 3.08 \\
0.8 & occupancy  & 29.1 &  82.7 & 2.46 & 40.2 & 110.1 & 3.24 \\
    & speed      & 29.5 &  84.7 & 2.69 & 28.5 &  92.7 & 2.60 \\
    & traveltime & 22.2 &  56.9 & 3.06 & 40.1 & 115.9 & 3.01 \\
1.0 & occupancy  & 28.4 &  81.8 & 2.99 & 28.2 &  90.1 & 2.96 \\
    & speed      & 28.4 &  81.8 & 2.99 & 28.1 &  90.1 & 2.96 \\
    & traveltime & 28.4 &  81.8 & 2.99 & 28.2 &  90.1 & 2.96 \\
\bottomrule
\end{tabular}
}
\end{table}


The best configuration for both strategies is $\alphaReg{=}0.2$ with the occupancy metric, yielding nearly identical MAE values (14.6 for simfree, 14.5 for simbased) but a lower RMSE for simfree (50.1 vs.\ 58.2). This dissociation indicates that by using the model calibrated with the simbased method occasionally, the augmentation procedure assigns a virtual sensor to an edge whose traffic profile departs significantly from the physical sensor, generating per-sensor spikes that inflate the RMSE while leaving the MAE largely unaffected. With the simfree model, by contrast, the augmentation method produces a more homogeneous error distribution. This difference is attributable to the spatial distribution of demand generated by the two calibration strategies.

Across all $\alphaReg$ values and metrics, error magnitudes are comparable in order of magnitude, proving that the method is not highly sensitive to the choice of simulation metric. The exception is $\alphaReg{=}0.2$ with occupancy, which achieves the lowest errors for both strategies. The improvement over pure turn-count scoring ($\alphaReg{=}1.0$, which yields RMSE of 81.8 for simfree) shows that a moderate amount of metric regularization is beneficial. However, a regularization factor $\alphaReg \geq 0.4$ degrades performance by pulling virtual sensors towards edges that are metrically similar to the physical sensor but volumetrically mismatched. The average hop count proves this interpretation: the best  configuration ($\alphaReg{=}0.2$, occupancy) selects virtual sensors at 1.86--1.92 hops on average, the shortest across all configurations, indicating that the scoring function converges to near-neighbor placements when both criteria agree.

All subsequent results are reported for the best configuration: hop range $[1,\,5]$, $\alphaReg{=}0.2$, occupancy metric.

\paragraph{Hourly vehicle count profiles.}

\begin{figure}[!ht]
    \centering
    \includegraphics[width=\linewidth]{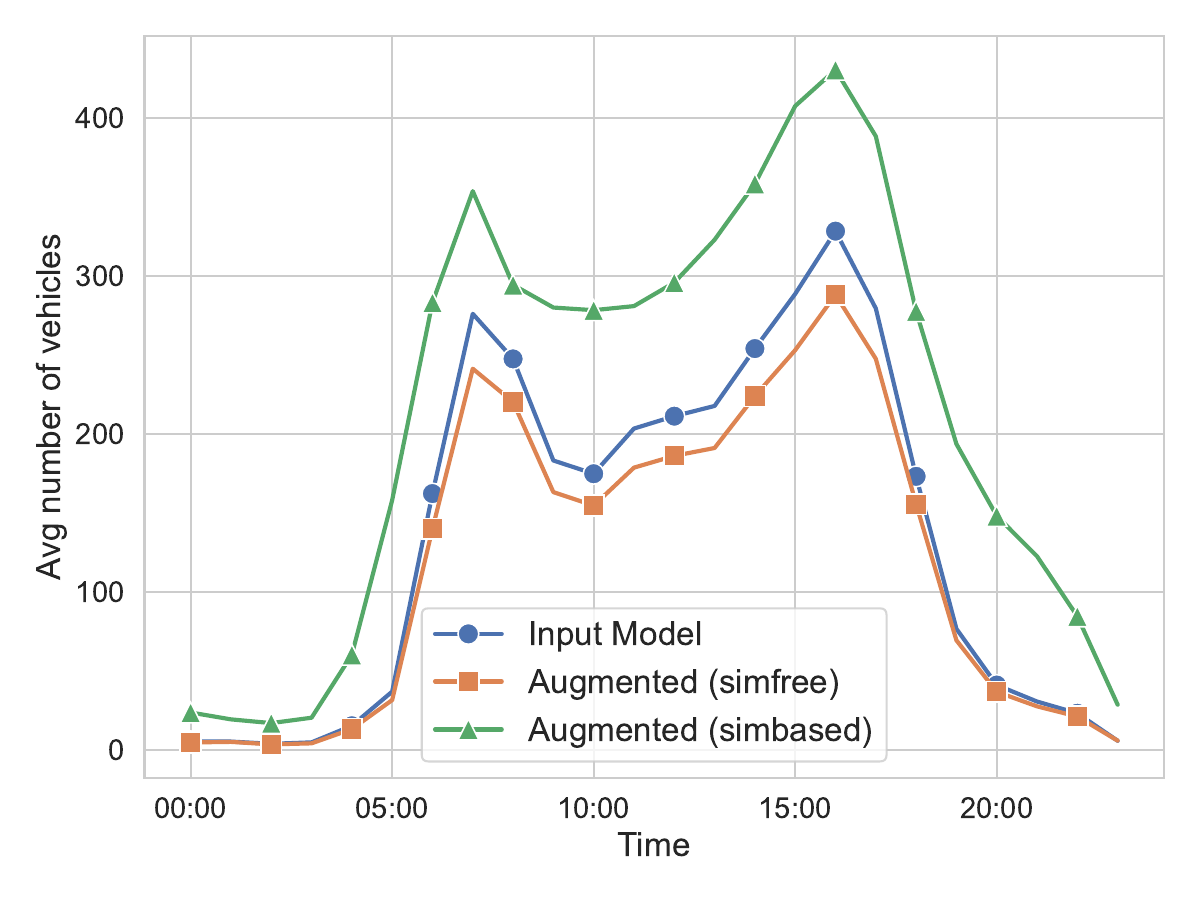}
    \caption{Average hourly vehicle count across all sensors for the input ground-truth, the simfree, and the simbased augmented models, at the best configuration ($\alphaReg{=}0.2$, occupancy, hop range $[1,\,5]$). Both augmented profiles reproduce the characteristic bimodal demand pattern of Brussels, with morning and evening peaks.}
    \label{fig:total_profile}
\end{figure}

Figure~\ref{fig:total_profile} compares the total hourly vehicle count summed across all sensors for the ground-truth, the augmented counts obtained using the simfree and the simbased augmented model.

Both augmented profiles reproduce the bimodal shape of the Brussels demand, with peaks during morning (07:00--09:00) and evening (17:00--19:00) rush hours, confirming that the temporal structure of urban mobility is preserved by the augmentation procedure. The augmented traffic counts generated using the simbased model tracks the aggregate ground-truth count more closely throughout the day, whereas the simfree model slightly underestimates the total volume during peak hours. However, the higher RMSE of the simbased model in Table~\ref{tab:augmentation_results} reveals that this aggregate accuracy is partly misleading: individual sensor errors partially cancel out when summed across the network, masking localised mismatches that remain visible at the per-sensor level. 

\paragraph{Sensor-level count accuracy.}

\begin{figure}[!ht]
    \centering
    \includegraphics[width=\linewidth]{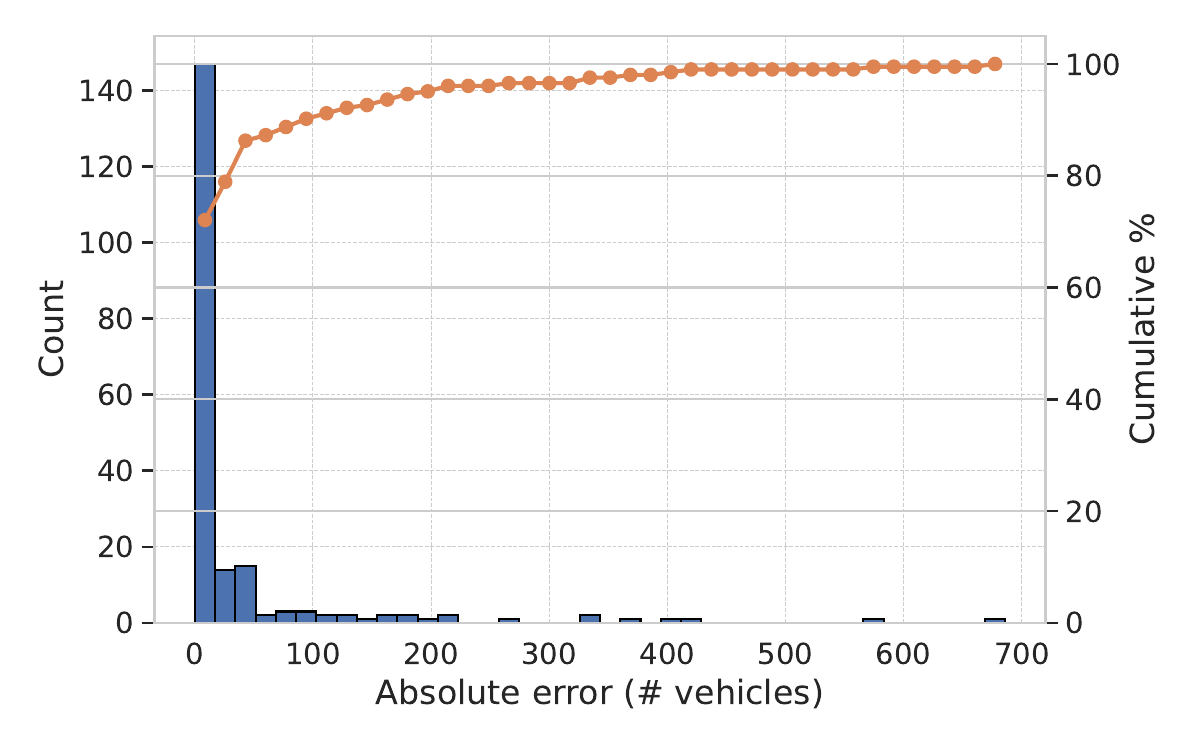}
    \caption{Distribution and cumulative percentage of the absolute count error per sensor (in vehicles per hour) between the augmented (simfree) and ground-truth datasets. The majority of sensor--interval pairs exhibit errors below 20 vehicles.}
    \label{fig:pareto}
\end{figure}

Figure~\ref{fig:pareto} shows the distribution of absolute per-sensor count error between the augmented (simfree, hop range [1,5], $\alphaReg{=}0.2$, occupancy metric) and the ground-truth models, aggregated across all sensors and all hourly intervals.

The majority of sensors exhibit errors below 20 vehicles per hour, and approximately 65\% of all sensor--interval pairs fall within this range, as indicated by the cumulative curve. The distribution is heavily right-skewed: a long tail extending to 700 vehicles per hour corresponds to a small number of sensors whose virtual counterparts are placed on edges with structurally different demand patterns, typically at the boundary of the road network where the candidate pool is smallest. These boundary cases are the primary drivers of the RMSE inflation observed for the simbased model.  

\paragraph{Spatiotemporal error structure.}

\begin{figure}[!ht]
    \centering
    \includegraphics[width=\linewidth]{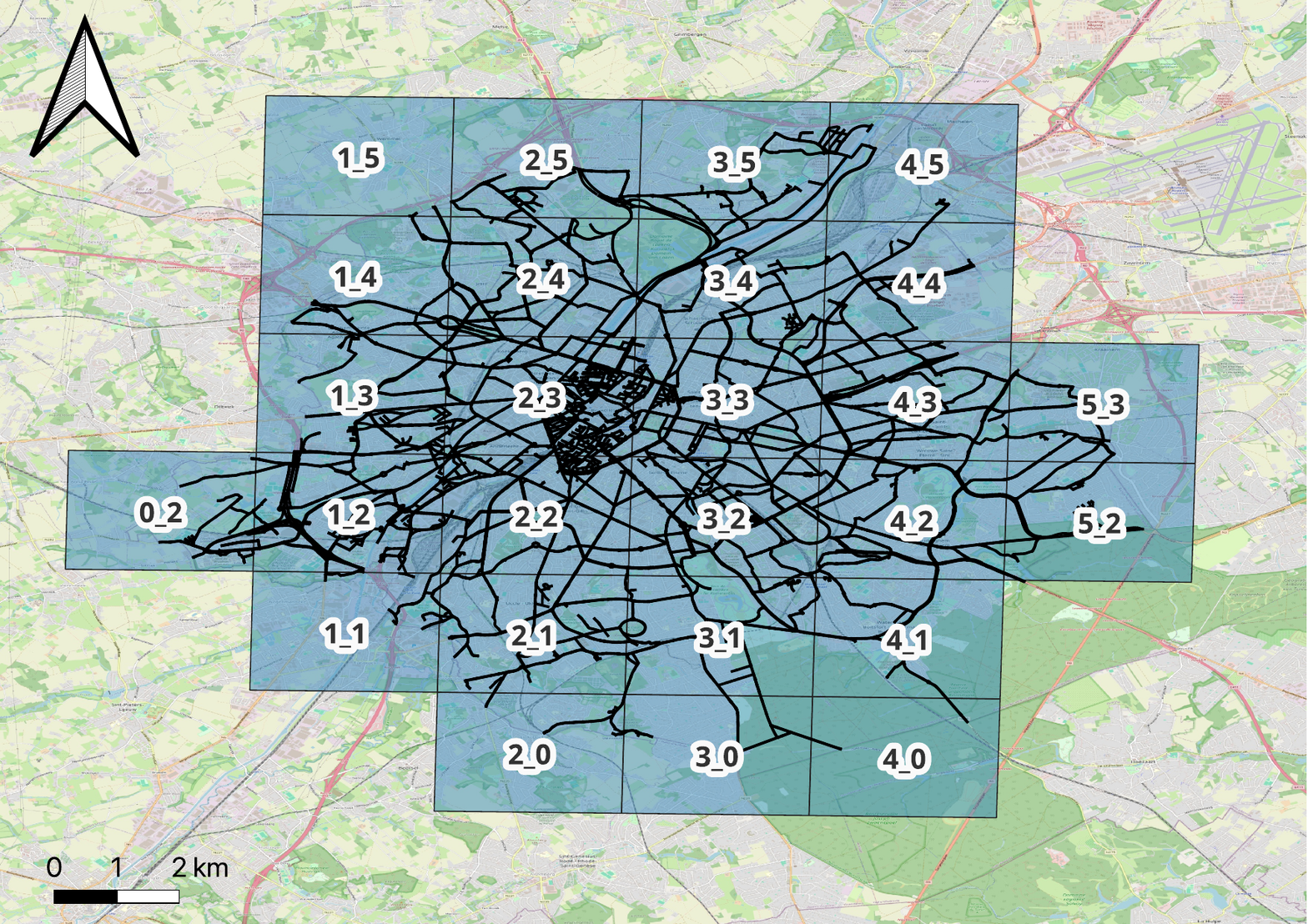}
    \caption{Spatial regions of $3000\,\text{m}^2$ used for the spatiotemporal
    error analysis.}
    \label{fig:bxl_regions}
\end{figure}

\begin{figure}[!ht]
    \centering
    \includegraphics[width=\linewidth]{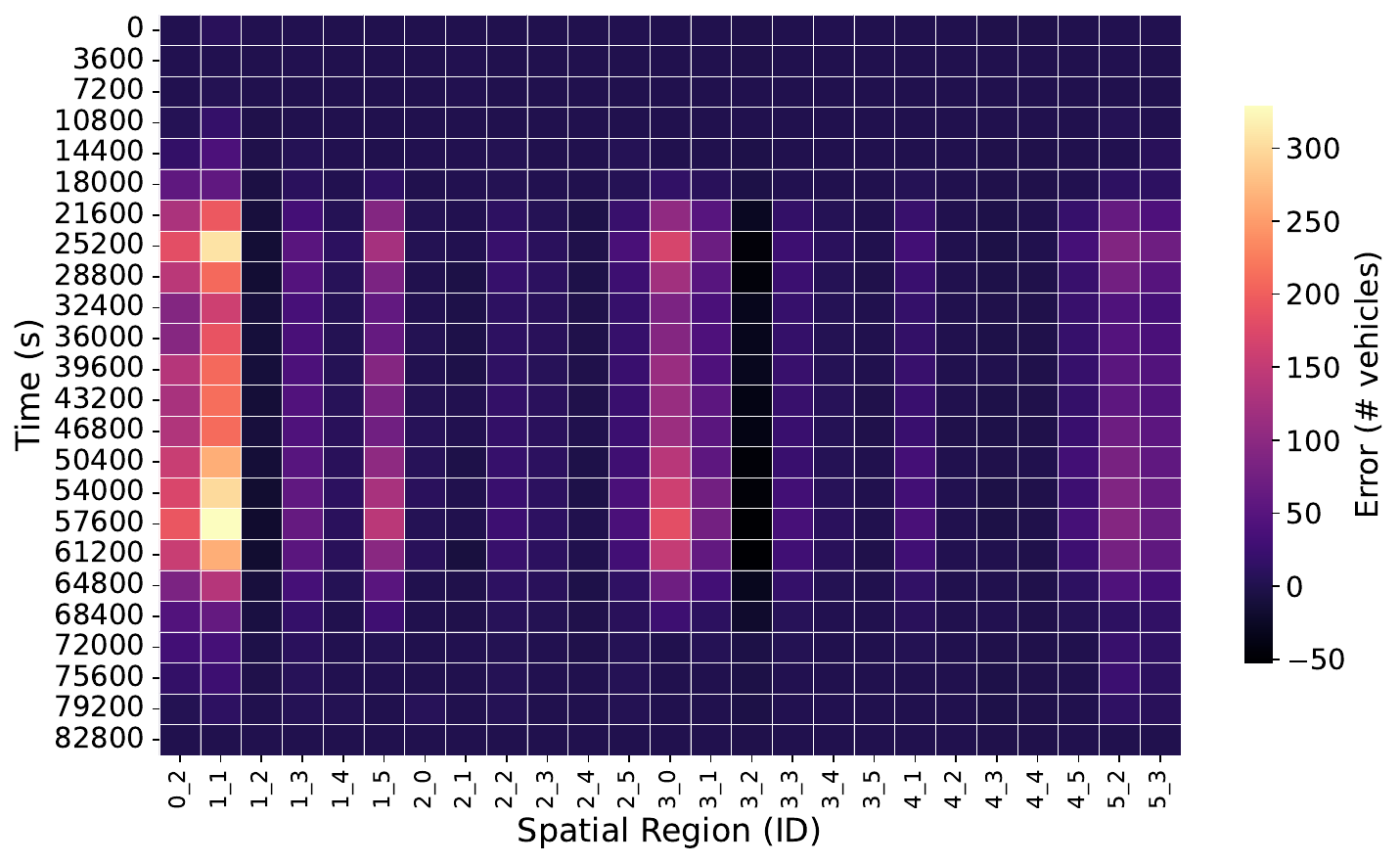}
    \caption{Signed error (augmented $-$ real) per spatial region and hour of day for the best simfree configuration ($\alphaReg{=}0.2$, occupancy, hop range $[1,\,5]$) for Brussels scenario. Negative values (dark blue) indicate underestimation; positive values (red to yellow) indicate overestimation.}
    \label{fig:heatmap}
\end{figure}

Figure~\ref{fig:heatmap} shows the signed error (augmented minus real) per spatial region and hour of day for the best simfree configuration (hop range [1,5], $\alphaReg{=}0.2$, occupancy metric). The modeled environment is partitioned into non-overlapping square regions of $3000~\text{m}^2$, following the same spatial decomposition used during calibration; each region aggregates the counts of all sensors (real or virtual) that fall within its boundaries, identified by a grid coordinate label as shown in Figure~\ref{fig:bxl_regions}. The granularity of the regions was selected as a compromise between spatial detail and model complexity, as this scale is consistent with ranges commonly used in urban transport-demand modeling and is sufficiently fine to capture variations in land use, population, and accessibility while avoiding an excessive number of zones~\cite{Altan2018_nd}.

The large majority of region--hour cells are near zero, indicating that the augmented dataset captures the spatial distribution of traffic for most of the network throughout the day. Deviations are concentrated in a small number of regions during the morning (06:00--10:00) and evening (15:00--19:00) peak hours: some regions show persistent underestimation (blue cells), while adjacent regions show symmetric overestimation (red cells). This complementary pattern is consistent with the heuristic misplacing a small number of virtual sensors to the wrong side of a congested junction, causing the demand to appear shifted spatially. Moreover, higher errors are observed mostly near the boundaries of the road network. In these areas, the reduced number of reachable candidate edges limits the ability of the matching procedure to identify suitable surrogate locations. This suggests that augmentation quality is partially influenced by local network topology and candidate availability.

\paragraph{Speed profile preservation.}

\begin{figure}[!ht]
    \centering
    \includegraphics[width=\linewidth]{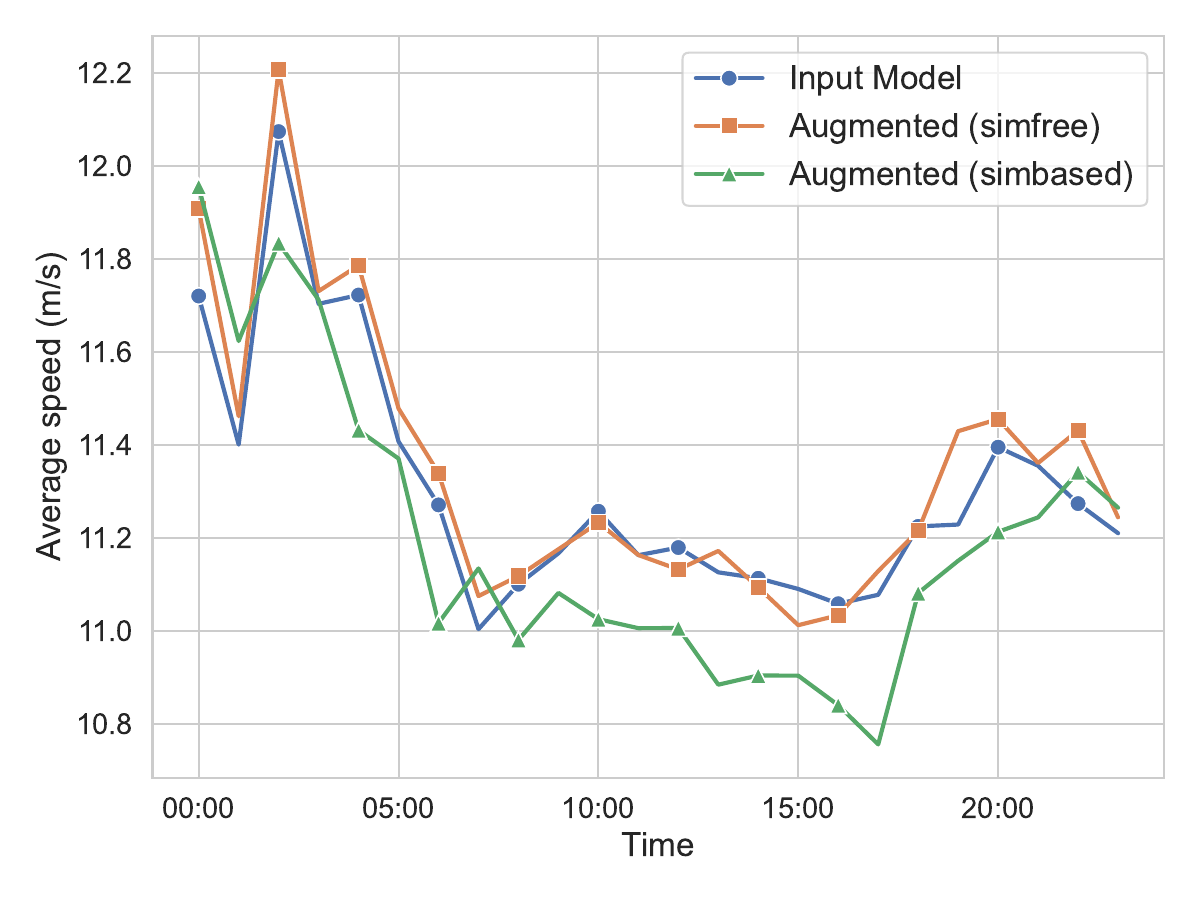}
    \caption{Network-wide average speed (m/s) per hour for the input ground-truth, simfree and simbased augmented models for Brussels scenario.}
    \label{fig:speed}
\end{figure}

Figure~\ref{fig:speed} compares the network-wide average speed per hour (in m/s) between the ground-truth and the augmented models, using the best configuration (hop range [1,5], $\alphaReg{=}0.2$, occupancy metric).

The profiles from input and augmented models are closely aligned across the full 24-hour period, including the characteristic speed drops during the morning and evening congestion peaks. This agreement proves that the virtual sensors are placed not only on edges with similar vehicle volumes, but also on edges where the underlying traffic dynamics are consistent with those at the original sensor locations. 

\subsubsection{Baseline Comparison}\label{sec:baseline}

To assess the contribution of the proposed scoring function, we compare $\mappingRealVirtual$ against two baseline placement strategies that operate over the same candidate pool $\candidates(t, \edge)$ produced by Algorithm~\ref{alg:get_candidates}, ensuring that any difference in outcome is attributable solely to the assignment criterion rather than to the set of reachable edges. 

The \emph{random placement} baseline assigns each sensored edge $\edge \in \edgesWSensors$ a virtual edge drawn uniformly at random from $\bigcup_{t} \candidates(t, \edge) \setminus \edgesWSensors$, with injectivity enforced by the same greedy matching used in Algorithm~\ref{alg:augmentation}. This baseline establishes a lower bound, testing whether any structured placement criterion adds value over chance.

The \emph{QR-pivot placement} baseline ranks all candidate edges globally by running a column-pivoted QR decomposition on the matrix $M \in \mathbb{R}^{\nintervals \times |\mathcal{E}_c|}$, where each column contains the hourly vehicle-count profile of one candidate sensored edge $\edgePrime \in \mathcal{E}_c = \bigcup_{\edge \in \edgesWSensors} \bigcup_{t} \candidates(t,\edge)$. The pivot order identifies the edges whose count profiles are most linearly independent, thereby maximizing the informational diversity of the augmented dataset. 
While the proposed method emphasizes similarity and consistency with the original sensor network, QR-pivot favors informational diversity by selecting traffic profiles that are as distinct as possible. Comparing the two approaches highlights the trade-off between preserving existing traffic dynamics and maximizing coverage of different traffic behaviors.


Table~\ref{tab:rmse_baseline} reports RMSE for the proposed method and both baselines across all hop ranges and calibration strategies.

\begin{table}[!ht]
\centering
\caption{RMSE for the proposed method and baseline placement strategies, across hop ranges and calibration strategies. Lower is better. The proposed method consistently outperforms both baselines for the two best-performing hop ranges $[1,\,5]$ and $[1,\,10]$. Note that in Proposed column we refer to the results of the best configuration (hop range [1,5], $\alphaReg{=}0.2$, occupancy metric).}
\label{tab:rmse_baseline}
\setlength{\tabcolsep}{8pt}
\renewcommand{\arraystretch}{1}
\small
\resizebox{\columnwidth}{!}{%
\begin{tabular}{l l r r r}
\toprule
\textbf{Calibration} & \textbf{Hop Range}
  & \textbf{Proposed} & \textbf{QR-Pivot} & \textbf{Random} \\
\midrule
\multirow{3}{*}{\textbf{Simfree}}
  & [1,5]   &  70.34 & 172.43 & 170.99 \\
  & [1,10]  &  96.87 & 173.72 & 181.45 \\
  & [3,10]  & 101.46 &  87.67 &  83.48 \\
\addlinespace
\multirow{3}{*}{\textbf{Simbased}}
  & [1,5]   &  96.81 & 185.11 & 192.63 \\
  & [1,10]  & 131.19 & 191.42 & 187.18 \\
  & [3,10]  & 138.83 & 173.81 & 147.45 \\
\bottomrule
\end{tabular}%
}
\end{table}

For the recommended hop range $[1,\,5]$, the proposed method outperforms both baselines under both calibration strategies. With the simfree calibration, the proposed method reduces RMSE by 59.2\% relative to QR-pivot (70.34 vs.\ 172.43) and by 58.9\% relative to random placement (70.34 vs.\ 170.99). Analogous improvements are observed for the simbased calibration (RMSE 96.81 vs.\ 185.11 for QR-pivot and 192.63 for random). Similar advantages hold for $[1,\,10]$, confirming that the similarity-based scoring function provides a consistent benefit when the candidate pool is non-trivially large.

The $[3,\,10]$ configuration is the only exception: here, both baselines outperform the proposed method. This reversal is explained by the structural mismatch between the scoring function and the enforced hop constraint. By excluding hops 1--2, the method is forced to select from edges that are topologically distant from the physical sensor; the turn-count and traffic-metric similarity between sensor and candidate are then inherently low, providing weak signal to the scoring function. Under these conditions, random and QR-pivot, which make no assumptions about similarity, can accidentally select near-neighbor edges reached via indirect paths, and the lack of meaningful signal in the scores makes the proposed method unable to distinguish good from poor candidates. This further motivates the adoption of $[1,\,5]$ as the recommended hop range for the considered scenario.

The results reveal a trade-off between accuracy and spatial diversity: restricting candidate sensors to nearby locations improves agreement with the ground-truth observations, whereas larger hop ranges increase spatial displacement at the cost of higher reconstruction errors. The proposed method therefore balances between preserving traffic characteristics and extending the spatial coverage of the dataset.

\subsection{Experimental Results (Namur Scenario)}

We further evaluate the proposed method on synthetic traffic models generated for the Namur road network (Figure~\ref{fig:namur_net}). In contrast to the Brussels case, where a calibrated model based on real sensor data is used as the ground-truth, the Namur experiments rely entirely on synthetic demand. This setup provides full control over traffic conditions and enables a statistical evaluation of augmentation quality across repeated experiments.

To assess robustness under different congestion levels, three demand scenarios are considered: 50,000, 100,000, and 150,000 vehicles distributed over a 24-hour period. For each demand level and for every combination of $\alphaReg$ and simulation metric, 30 independent simulations are performed. All simulations use the same set of virtual sensors, covering approximately 30\% of the road network. The augmented dataset is generated by applying the proposed heuristic to the corresponding synthetic model, and the reported results are averaged over all simulations for each demand level.

Since the experiments involve different traffic demand levels, we employ the Relative Mean Absolute Error (RMAE) as a scale-independent accuracy metric:

\begin{equation}
  \textnormal{RMAE}(\hat{y},y) =
    \frac{\sum_{i=1}^{n} |y_i - \hat{y}_i|}{\sum_{i=1}^{n} y_i}
\end{equation}

\noindent where $n$ is the number of sensor--hour observations, and $\simulator$ and $\avgRealTraffic$ are the augmented and real traffic counts respectively.

Table~\ref{tab:namur_errors} reports the RMAE averaged over the 30 experiments. The hop range $[1,\,5]$ consistently yields the lowest RMAE across all demand levels. The degradation at $[3,\,10]$ indicates that excluding the nearest candidates removes the most similar edges from the pool, forcing the method to select from edges that share little traffic continuity with the physical sensor, and RMAE more than triples. The standard deviations across the 30 runs are consistently low for $[1,\,5]$, proving that the heuristic converges to solutions of comparable quality regardless of the specific sensor configuration realized in each run.

\begin{table}[!ht]
\centering
\caption{Best RMAE (mean $\pm$ std, lower is better) for each demand level and augmentation metric across the three evaluation windows (hop interval $h_1$--$h_2$) for the Namur scenario, with the optimal $\alpha$ in parentheses. Bold highlights the best result per row.}
\label{tab:namur_errors}
\setlength{\tabcolsep}{6pt}
\renewcommand{\arraystretch}{1.15}
\small
\resizebox{\columnwidth}{!}{%
\begin{tabular}{lllll}
\toprule
Demand & Metric & $h_1$--$h_2$: 1--5 & $h_1$--$h_2$: 1--10 & $h_1$--$h_2$: 3--10 \\
\midrule
\multirow{3}{*}{50k} & Occupancy & \textbf{$0.296 \pm 0.017$ ($\alpha$=0.2)} & $0.521 \pm 0.020$ ($\alpha$=0.2) & $0.753 \pm 0.109$ ($\alpha$=0.0) \\
  & Speed & \textbf{$0.231 \pm 0.020$ ($\alpha$=0.2)} & $0.640 \pm 0.059$ ($\alpha$=0.2) & $0.660 \pm 0.013$ ($\alpha$=0.0) \\
  & Travel time & \textbf{$0.558 \pm 0.084$ ($\alpha$=1.0)} & $0.739 \pm 0.109$ ($\alpha$=0.0) & $0.815 \pm 0.111$ ($\alpha$=0.0) \\
\cmidrule(lr){2-4}
\multirow{3}{*}{100k} & Occupancy & \textbf{$0.287 \pm 0.023$ ($\alpha$=0.2)} & $0.521 \pm 0.056$ ($\alpha$=0.2) & $0.681 \pm 0.076$ ($\alpha$=0.0) \\
  & Speed & \textbf{$0.228 \pm 0.041$ ($\alpha$=0.2)} & $0.604 \pm 0.068$ ($\alpha$=0.2) & $0.661 \pm 0.023$ ($\alpha$=0.0) \\
  & Travel time & \textbf{$0.521 \pm 0.043$ ($\alpha$=1.0)} & $0.728 \pm 0.147$ ($\alpha$=0.0) & $0.816 \pm 0.122$ ($\alpha$=0.0) \\
\cmidrule(lr){2-4}
\multirow{3}{*}{150k} & Occupancy & \textbf{$0.311 \pm 0.076$ ($\alpha$=0.2)} & $0.523 \pm 0.165$ ($\alpha$=0.2) & $0.712 \pm 0.099$ ($\alpha$=0.0) \\
  & Speed & \textbf{$0.300 \pm 0.082$ ($\alpha$=0.2)} & $0.677 \pm 0.052$ ($\alpha$=0.0) & $0.682 \pm 0.052$ ($\alpha$=0.0) \\
  & Travel time & \textbf{$0.506 \pm 0.043$ ($\alpha$=0.2)} & $0.779 \pm 0.117$ ($\alpha$=0.0) & $0.826 \pm 0.146$ ($\alpha$=0.0) \\
\cmidrule(lr){2-4}
\bottomrule
\end{tabular}
}
\end{table}

Across demand levels, RMAE remains stable for $[1,\,5]$, indicating that the method scales well from moderate to high congestion regimes. 



\begin{figure}[!ht]
    \centering
    \includegraphics[width=\linewidth]{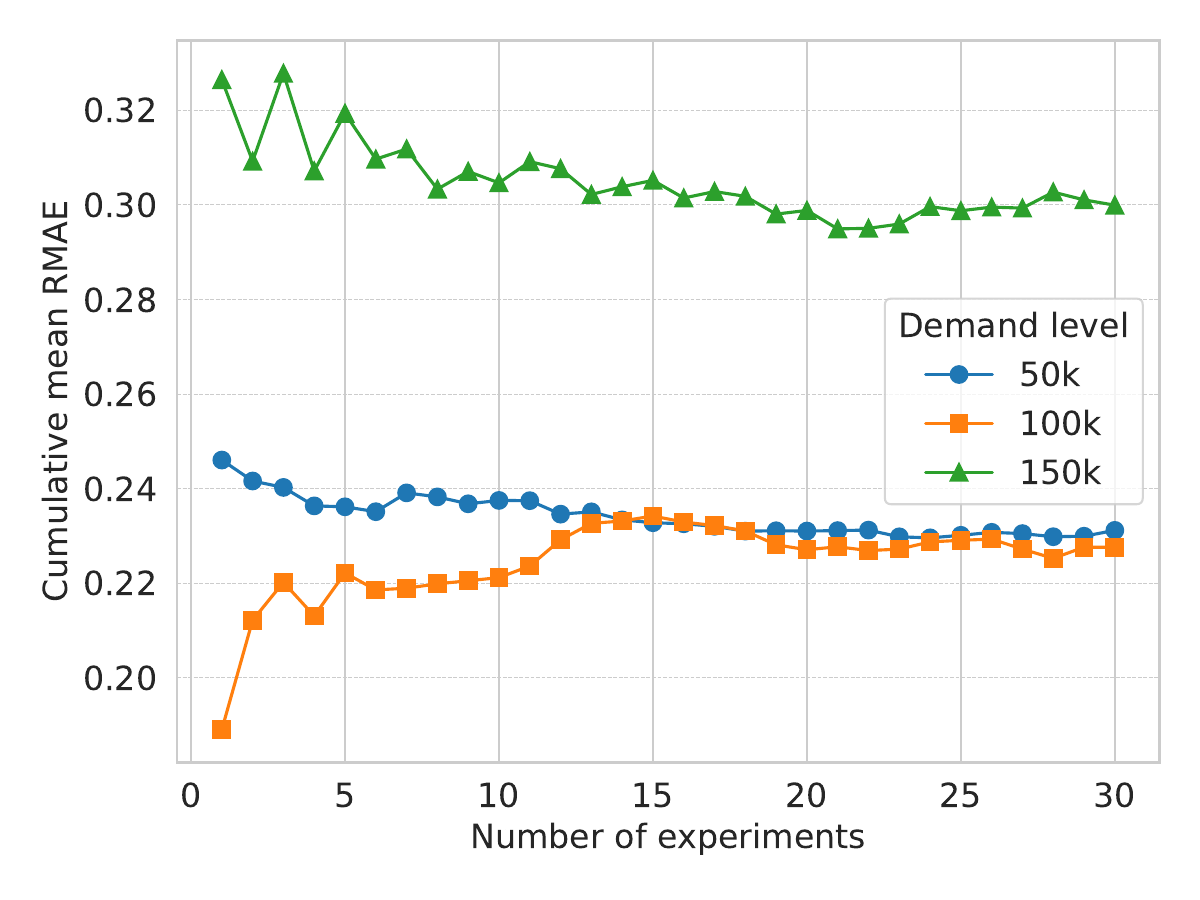}
    \caption{Cumulative mean RMAE across the 30 independent runs for the three demand levels (50k, 100k, 150k vehicles).}
    \label{fig:cum_rmae_namur}
\end{figure}

Figure~\ref{fig:cum_rmae_namur} shows the cumulative mean RMAE across 30 experiments for each demand level, computed over all runs corresponding to the best configuration ($\alphaReg$, metric) identified for that level.  The cumulative mean stabilizes rapidly, within the first 15 experiments, and remains nearly constant across all three demand levels thereafter. The RMAE is highest for the 150000 vehicle demand scenario, which is expected: at higher demand levels, greater traffic complexity increases uncertainty in flow observations, leading to a less precise virtual sensor allocation. Even in this case, however, the cumulative mean stabilizes after approximately 20 experiments. Beyond that point, only minor fluctuations are visible for all demand levels, indicating that 30 runs constitute a statistically sufficient sample to evaluate the accuracy of the proposed method.

\section{Discussion}\label{sec:discussion}

The experimental results prove that the proposed simulation-based heuristic reliably generates augmented traffic datasets that preserve the spatio-temporal structure of urban mobility at both the sensor and the regional level. The key conditions for this to hold are: (i) a hop range that includes near-neighbor candidates ($\hmin{=}1$, $\hmax{\leq}5$ recommended), (ii) a small amount of metric regularization ($\alphaReg{=}0.2$) to refine turn-count matching without distorting it, and (iii) a calibration strategy that produces a spatially smooth demand distribution. When these conditions are met, the method outperforms both random placement and diversity-maximizing QR-pivot, proving that turn-count continuity and traffic-metric similarity are effective proxies for sensor surrogacy.

Several limitations of the current approach should be acknowledged. First, the quality of the augmented dataset is bounded by the quality of the underlying traffic model. Since virtual sensor observations are generated from simulated traffic flows, calibration errors may propagate to the augmented data. Consequently, the proposed method should be viewed as a mechanism for extending existing traffic information rather than correcting inaccuracies in the input model. Traffic calibration is an underdetermined problem: for a given set of sensor observations, many distinct parameter configurations can produce equally plausible traffic flows. A calibrated model may match aggregate counts at instrumented locations while assigning vehicles to routes that differ substantially from their real trajectories, and virtual sensor readings inherit this ambiguity directly. Reducing this gap remains an open challenge and is tightly coupled to advances in traffic calibration methodology and sensor density.

Second, we do not model realistic traffic light programs due to the absence of publicly available signal data, and we do not apply corrections for OSM-to-SUMO conversion errors. These simplifications affect the realism of the simulation model but do not undermine the primary validation goal, which is to demonstrate that input and augmented datasets lead to similar traffic patterns when evaluated on disjoint sets of sensors.

Third, while the minimum hop distance $\hmin$ provides a configurable mechanism to control the spatial displacement between physical and virtual sensors, a formal privacy analysis quantifying how difficult it is to reconstruct individual vehicle trajectories from the augmented data, should be considered.

A seemingly circular objection to the proposed method is that, since the input traffic model is itself calibrated from real sensor data, the augmentation process cannot produce information beyond what is already contained in the original sparse observations. Calibration estimates a set of vehicle trajectories such that the simulated and the real counts at sensors location are similar. Data augmentation, as proposed here, enables extracting additional virtual observations from traffic models (through simulation) with no additional cost. In other words, calibration is where the (limited) real data is used to constrain the model; augmentation only exploits information that the calibration step has already inferred, structured, and validated against the observed counts. That is, it does not introduce new real-world information, but makes the already-inferred network-wide state usable as a richer training/evaluation dataset. 
\section{Conclusion and Future Work}\label{sec:conclusion}

This paper presents a novel simulation-based heuristic for vehicular traffic data augmentation to address the challenges of data scarcity. The goal of this work is to establish the augmentation methodology itself and to evaluate its ability to preserve traffic dynamics under spatial sensor displacement.

The proposed heuristic exploits simulation-derived turn counts and traffic metrics to score candidate virtual sensor locations and selects placements via a greedy injective matching. The method is computationally lightweight, requires no training data, and produces a deterministic output (given $\epsExplore{=}0$) that can be reproduced from any calibrated or synthetic simulation model. The experiments on the calibrated Brussels network and the synthetic Namur scenarios prove that the augmented datasets replicate the spatio-temporal patterns of the original traffic counts at both the sensor level and the spatial-region level. The baseline comparison shows that the proposed scoring function reduces RMSE by approximately 59\% relative to random and QR-pivot placement for the recommended $[1,\,5]$ hop range, proving that turn-count continuity and traffic-metric similarity are more effective placement criteria than chance or diversity maximization.

Investigating the impact of the generated datasets on learning, forecasting, and control tasks will be addressed in future work. Furthermore, future work will focus on three directions. First, we will investigate alternative scoring formulations and search strategies to improve placement quality in sparse regions and under wider hop constraints. Second, we will explore the joint use of real and augmented data to potentially enhance the accuracy of downstream predictive traffic models. Third, we will conduct a formal privacy analysis to quantify how difficult it is to reconstruct individual vehicle trajectories from the augmented dataset, providing a rigorous characterization of the privacy--utility trade-off that the hop constraint is designed to control.

Beyond extending sensor coverage, the proposed method could support policy-oriented analyses on corridors lacking direct instrumentation, such as estimating the CO$_2$ impact of converting a traffic lane into a dedicated bus lane, or assessing how displaced private-vehicle traffic affects the reliability of nearby bus routes. By placing virtual sensors on such corridors and extracting traffic metrics from the simulation before and after a proposed intervention, the augmented dataset could feed standard emission models or multimodal transit analyses. Validating virtual sensor reliability under such counterfactual network configurations is left for future work.



\printbibliography


\end{document}